%% file: dd-transformer-coling2020.tex
\documentclass[11pt]{article}
\usepackage{coling2020}
\usepackage{times}
\usepackage{url}
\usepackage{latexsym}

\usepackage[T2A,T1]{fontenc} %
\usepackage[utf8]{inputenc}
\usepackage[russian,english]{babel}

\usepackage{soul}
\usepackage{graphicx}
\usepackage{subcaption}
\usepackage{comment}
\usepackage{booktabs}
\usepackage[table, dvipsnames]{xcolor}
\usepackage{calc}
\usepackage{caption}
\usepackage{varwidth}
\usepackage{multicol}
\usepackage{multirow}

\usepackage{tikz}
\usetikzlibrary{backgrounds,calc,shadings,shapes.arrows,arrows,shapes.symbols,shadows,positioning,decorations.markings,backgrounds,arrows.meta,shapes.geometric,patterns}
\tikzstyle{input} = [rectangle, rounded corners, minimum width=1cm, minimum height=1.2cm,text centered, align=center, draw=white, fill=white]
\tikzstyle{output} = [rectangle, rounded corners, minimum width=1cm, minimum height=1.2cm,text centered, align=left, draw=white, fill=white]
\tikzstyle{arrow} = [thick,->,>=stealth]

\usepackage{pgfplots}
\DeclareUnicodeCharacter{2212}{−}
\usepgfplotslibrary{groupplots,dateplot}
\pgfplotsset{compat=newest}

\usepackage{siunitx}
\definecolor{MyBlue}{rgb}{0.0, 0.18, 0.65}
\colorlet{HighlightColor}{MyBlue}
\definecolor{applegreen}{rgb}{0.55, 0.71, 0.0}

\usepackage{xspace}
\makeatletter
\DeclareRobustCommand\onedot{\futurelet\@let@token\@onedot}
\def\@onedot{\ifx\@let@token.\else.\null\fi\xspace}
\def\eg{\emph{e.g}\onedot} 
\def\ie{\emph{i.e}\onedot} 
\def\cf{\emph{c.f}\onedot} 
\def\etc{\emph{etc}\onedot} \def\vs{\emph{vs}\onedot}

\makeatother

\usepackage[font=small,labelfont=bf]{caption}

\usepackage[
    pagebackref=true,
    breaklinks=true,
    pdftitle={Dual-decoder Transformer for Joint Automatic Speech Recognition and Multilingual Speech Translation},
    pdfsubject={Natural Language Processing, Speech Recognition, Speech Translation},
    pdfkeywords={transformer, multilingual speech translation, speech recognition, machine translation}]{hyperref}

\input{math}

\colingfinalcopy %

\newcommand{\textsup}[1]{{\normalfont\normalsize\textsuperscript{#1}}} 
\newcommand{\var}[1]{\texttt{#1}}

\newcommand{\highlight}[1]{{\color{HighlightColor}\underline{#1}}}

\title{Dual-decoder Transformer for Joint Automatic Speech Recognition and Multilingual Speech Translation}

\author{Hang Le\textsup{1}\qquad Juan Pino\textsup{2} \qquad Changhan Wang\textsup{2}\\\bfseries\large
Jiatao Gu\textsup{2}\qquad Didier Schwab\textsup{1} \qquad Laurent Besacier\textsup{1}\\
{\normalsize\textsuperscript{1}Univ. Grenoble Alpes, CNRS, LIG \qquad \textsuperscript{2}Facebook AI}\\
{\small\tt \{hang.le, didier.schwab, laurent.besacier\}@univ-grenoble-alpes.fr}\\
{\small\tt \{juancarabina, changhan, jgu\}@fb.com}
  }

\date{}

\begin{document}
\maketitle
\begin{abstract}
We introduce dual-decoder Transformer, a new model architecture that jointly performs automatic speech recognition (ASR) and multilingual speech translation (ST). Our models are based on the original Transformer architecture~\cite{vaswani2017attention} but consist of two decoders, each responsible for one task (ASR or ST). Our major contribution lies in how these decoders interact with each other: one decoder can attend to different information sources from the other via a \emph{dual-attention} mechanism. We propose two variants of these architectures corresponding to two different levels of dependencies between the decoders, called the \emph{parallel} and \emph{cross} dual-decoder Transformers, respectively. 
Extensive experiments on the MuST-C dataset show that our models outperform the previously-reported highest translation performance in the multilingual settings, and outperform as well bilingual one-to-one results. Furthermore, our parallel models demonstrate no trade-off between ASR and ST compared to the vanilla multi-task architecture. Our code and pre-trained models are available at \url{https://github.com/formiel/speech-translation}.
\end{abstract}

\section{Introduction}\label{sec:introduction}

While cascade speech-to-text translation (ST) 
systems operate in two steps: source language automatic speech recognition (ASR) and source-to-target text machine translation (MT), recent works have attempted to build end-to-end ST without using source language transcription during decoding \cite{berard-nips2016,weiss2017sequence,DBLP:journals/corr/abs-1802-04200}. After two years of extensions to these pioneering works, the last results of the IWSLT 2020 shared task on offline speech translation \cite{ansari-etal-2020-findings} demonstrate that end-to-end models are now on par (if not better) than their cascade counterparts.  %
Such a finding motivates even more strongly the works on multilingual (one-to-many, many-to-one, many-to-many) ST \cite{di2019one,inaguma2019multilingual,wang2020covost} for which end-to-end models are well adapted by design.
Moreover, of these two approaches: \textit{cascade} proposes a very loose integration of ASR and MT (even if lattices or word confusion networks were used between ASR and MT before end-to-end models appeared) while most \textit{end-to-end} approaches simply ignore ASR subtask, trying to directly translate from source speech to target text. We believe that these are two edge design choices and that a tighter coupling of ASR and MT is desirable %
for future end-to-end ST applications, in which the display of transcripts alongside translations can be beneficial to the users~\cite{sperber2020consistent}.

This paper addresses multilingual ST and investigates more closely the interactions between speech transcription (ASR) and speech translation (ST) in a multilingual end-to-end architecture based on Transformer. While those interactions were previously investigated as a simple multi-task framework for a bilingual case~\cite{anastasopoulos2018tied}, we propose a dual-decoder with an ASR decoder tightly coupled with an ST decoder and evaluate its effectiveness on one-to-many ST. Our model is inspired by \newcite{liu2019synchronous}, but the interaction between ASR and ST decoders is much tighter.\footnote{The model of \newcite{liu2019synchronous} does not have interaction between internal hidden states of the decoders (\cf Section~\ref{sec:dd-variants}).}
Finally, experiments show that our model outperforms theirs on the MuST-C benchmark~\cite{di2019must}.

Our contributions are summarized as follows:
(1) a new model architecture for joint ASR and multilingual ST;
(2) an integrated beam search decoding strategy which jointly transcribes and translates, and that is extended to a wait-$k$ strategy where the ASR hypothesis is ahead of the ST hypothesis by $k$ tokens and vice-versa;
and (3) competitive performance on the MuST-C dataset 
in both bilingual and multilingual settings and improvements on previous joint ASR/ST work.

\section{Related Work}

\paragraph{Multilingual ST} 
Multilingual translation~\cite{googlemulti2017} consists in translating between different language pairs with a single model, thereby improving maintainability and the quality of low resource languages. 
\newcite{di2019one} adapt this method to one-to-many multilingual speech translation by adding a language embedding to each source feature vector. They also observe that using the source language (English) as one of the target languages improves performance. \newcite{inaguma2019multilingual} simplify the previous approach by prepending a target language token to the decoder and apply it to one-to-many and many-to-many speech translation. They do not investigate many-to-one due to the lack of a large corpus for this. To fill this void, \newcite{wang2020covost} release the CoVoST dataset for ST from 11 languages into English and demonstrate the effectiveness of many-to-one ST.

\paragraph{Joint ASR and ST} 
Joint ASR and ST decoding was first proposed by \newcite{anastasopoulos2018tied} through a multi-task learning framework. \newcite{chuang2020worse} improve multitask ST by using word embedding as an intermediate level instead of text. A two-stage model that performs first ASR and then passes the decoder states as input to a second ST model was also studied previously \cite{anastasopoulos2018tied,sperber2019attention}. This architecture is closer to cascaded translation while maintaining end-to-end trainability. \newcite{sperber2020consistent} introduce the notion of consistency between transcripts and translations and propose metrics to gauge it. They evaluate different model types for the joint ASR and ST task and conclude that end-to-end models with coupled inference procedure are able to achieve strong consistency. In addition to existing models having coupled architectures, they also investigate a model where the transcripts are concatenated to the translations, and the shared encoder-decoder network learns to predict this concatenated outputs. It should be noted that our models have lower latency compared to this approach since the concatenation of outputs makes the two tasks sequential in nature. Our work is closely related to that of \newcite{liu2019synchronous} who propose an \textit{interactive attention} mechanism which enables ASR and ST to be performed synchronously. Both ASR and ST decoders do not only rely on their previous outputs but also on the outputs predicted in the other task. We highlight three differences between their work and ours: (a) we propose a more general framework in which \cite{liu2019synchronous} is a special case; (b) tighter integration of ASR and ST is proposed in our work; and (c) we experiment in a multilingual ST setting while previous works on joint ASR and ST only investigated bilingual ST.

\section{Dual-decoder Transformer for Joint ASR and Multilingual ST}

We now  present the proposed dual-decoder Transformer for jointly performing ASR and multilingual ST. Our models are based on the Transformer architecture~\cite{vaswani2017attention} but consist of two decoders. Each decoder is responsible for one task (ASR or ST). The intuition is that the problem at hand consists in solving two different tasks with different characteristics and different levels of difficulty (multilingual ST is considered more difficult than ASR). Having different decoders specialized in different tasks may thus produce better results. In addition, since these two tasks can be complementary, it is natural to allow the decoders to help each other. Therefore, in our models, we introduce a \emph{dual-attention} mechanism: in addition to attending to the encoder, the decoders also attend to each other. 

\subsection{Model overview}

The model takes as input a sequence of speech features $\x=(x_1,x_2,\dots,x_{T_x})$ in a specific \emph{source} language (\eg English)
and outputs a transcription $\y = (y_0,y_1,\dots,y_{T_y})$ in the same language as well as translations $\z_{1}, \z_{2}, \dots, \z_{M}$ in $M$ different \emph{target} languages (\eg French, Spanish, \etc). When $M = 1$, this corresponds to joint ASR and bilingual ST~\cite{liu2019synchronous}. For simplicity, our presentation considers only a single target language with output $\z = (z_0,z_1,\dots,z_{T_z})$. All results, however, apply to the general multilingual case.
In the sequel, denote $\y_{< t} \triangleq (y_0,y_1,\dots,y_{t-1})$ and $\y_{> t} \triangleq (y_{t+1},y_{t+2},\dots,y_{T_y})$ ($y_t$ is included if ``$<$'' and ``$>$'' are replaced by ``$\le$'' and ``$\ge$'' respectively). In addition, assume that $y_t$ is ignored if $t$ is outside of the interval $[0, T_y]$. Notations apply to $\z$ as well.

The dual-decoder model jointly predicts the transcript and translation in an autoregressive fashion:
\begin{equation}\label{eq:joint-distribution}
p(\y,\z \mid \x)  = \prod_{t=0}^{\max(T_y,T_z)} p(y_t,z_t \mid \y_{< t}, \z_{< t}, \x).
\end{equation}
A natural model would consist of a single decoder followed by a softmax layer.
However, even if the capacity of the decoder were large enough for handling both ASR and ST generation,
a single softmax would require a very large joint vocabulary (with size $V_yV_z$ where $V_y$ and $V_z$ are respectively the vocabulary sizes for $\y$ and $\z$). Instead, our dual-decoder consists of two sub-decoders that are specialized in producing outputs tailored to the ASR and ST tasks separately.
Formally, our model predicts the next output tokens $(\hat{y}_s,\hat{z}_t)$ (where $1\le s \le T_y, 1\le t \le T_z$) given a pair of previous outputs $(\y_{< s},\z_{< t})$ as:
\begin{align}
\h^y_s, \h^z_t &= \decoder_\mr{dual}(\y_{< s},\z_{< t},\encoder(\x)) \in \RR^{d_y}\times \RR^{d_z}, \label{eq:dd-embedding}\\
p(y_s \mid \y_{< s},\z_{< t},\x) &= [\softmax(\W^y\h^y_{s} + \b^y)]_{y_s},\qquad \hat{y}_s = \mathrm{argmax}_{y_s}\ p(y_s \mid \y_{< s},\z_{< t},\x), \label{eq:dd-prediction-y}\\
p(z_t \mid \y_{< s},\z_{< t},\x) &= [\softmax(\W^z\h^z_{t} + \b^z)]_{z_t},\qquad\hspace{0.16em}
\hat{z}_t = \mathrm{argmax}_{z_t}\ p(z_t \mid \y_{< s},\z_{< t},\x), \label{eq:dd-prediction-z}
\end{align}
where $[\v]_i$ denotes the $i\textsuperscript{th}$ element of the vector $\v$. Note that $y_s$ and $z_t$ are token indices ($1\le y_s \le V_y, 1\le z_t \le V_z$).  In~\eqref{eq:dd-prediction-y} and~\eqref{eq:dd-prediction-z}, we detail the intermediate quantities $p(y_s \mid \cdot)$ and $p(z_t \mid \cdot)$ as obtained from the probability distributions over the output vocabulary.
In the above, we have made an important assumption about the joint probability $p(y_s,z_t \mid \cdot)$ that it can be factorized into $p(y_s \mid \cdot)p(z_t \mid \cdot)$.
Therefore, the joint distribution~\eqref{eq:joint-distribution} encoded by the dual-decoder Transformer can be rewritten as
\begin{equation}\label{eq:joint-distribution-factorized}
p(\y,\z \mid \x)  = \prod_{t=0}^{\max(T_y,T_z)} p(y_t \mid \y_{< t}, \z_{< t}, \x)p(z_t \mid \y_{< t}, \z_{< t}, \x).
\end{equation}
We also assumed so far that the sub-decoders start at the same time, which is the most basic configuration. In practice, however, one may allow one sequence to advance $k$ steps compared to the other, known as the \emph{wait-$k$} policy~\cite{Ma19acl}. %
For example, if ST waits for ASR to produce its first $k$ tokens, then the joint distribution becomes
\begin{align}
p(\y,\z \mid \x) &= p(\y_{< k} \mid \x) p(\y_{\ge k},\z \mid \y_{< k},\x) \\
&=  \prod_{t=0}^{k-1} p(y_t \mid y_{<t}, \x) \prod_{t=0}^{\max(T_y-k,T_z)} p(y_{t+k}\mid \y_{< t+k}, \z_{< t}, \x)p(z_{t}\mid \y_{< t+k}, \z_{< t}, \x). \label{eq:wait-k-distribution}
\end{align}

In the next section, we propose two concrete architectures for the dual-decoder, corresponding to different levels of  dependencies between the two sub-decoders (ASR and ST). Then, we show that several known models in the literature are special cases of these architectures.

\subsection{Parallel and cross dual-decoder Transformers}
\label{sec:dd_parallel}
\begin{figure}[!htb]
\makebox[\textwidth][c]{
    \centering
    \begin{minipage}[b]{0.773\linewidth}
	    \begin{subfigure}[b]{\linewidth}
	      \centering
	      \includegraphics[width=\linewidth]{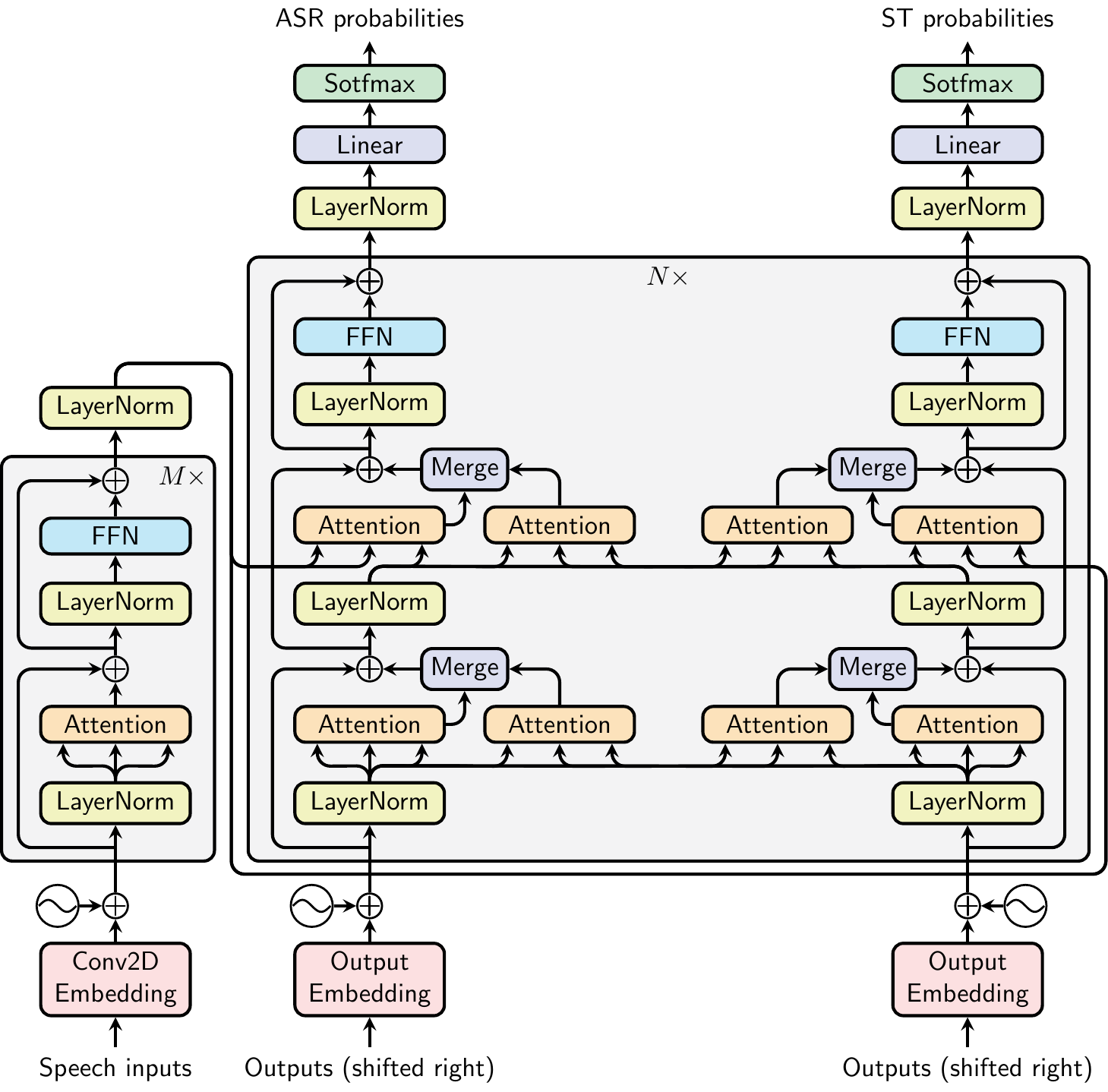}
	      \caption{The parallel dual-decoder Transformer}
	      \label{fig:dd-parallel}
	    \end{subfigure} 
	\end{minipage}
	\begin{minipage}[b]{0.255\linewidth}
		\begin{subfigure}[b]{\linewidth}
			\centering
			\includegraphics[width=\linewidth]{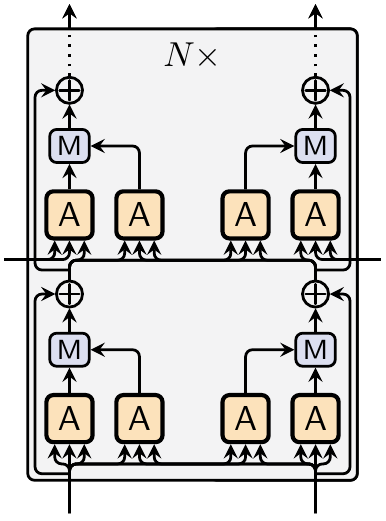}
			\caption{Parallel}
			\label{fig:dd-parallel-semantic}
		\end{subfigure}\\ \vspace*{0.1mm}
	    \begin{subfigure}[b]{\linewidth}
	      \centering
	      \includegraphics[width=\linewidth]{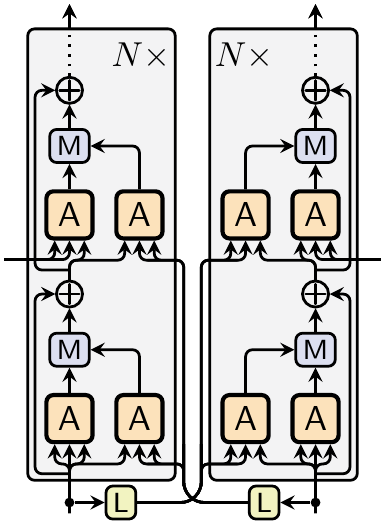}%
	      \caption{Cross}
	      \label{fig:dd-cross-semantic}
	    \end{subfigure}
    \end{minipage}
}
\caption{The dual-decoder Transformers. Figure (a) shows the detailed architecture of the \emph{parallel} dual-decoder Transformer, and Figure (b) shows its simplified view. The \emph{cross} dual-decoder Transformer is very similar to the parallel one, except that the keys and values fed to the dual-attention layers come from the previous output, which is illustrated by Figure (c). From the above figures, one can easily infer the detailed architecture of the cross Transformer, which can be found in the Appendix. 
Abbreviations: \textbf{A} (Attention), \textbf{M} (Merge),  \textbf{L} (LayerNorm). 
}
\label{fig:dd-transformer}
\end{figure}

The first architecture is called \emph{parallel dual-decoder Transformer}, which has the highest level of dependencies: one decoder uses the hidden states of the other to compute its outputs, as illustrated in Figure~\ref{fig:dd-parallel}.
The encoder consists of an input embedding layer followed by a positional embedding and a number of \emph{self-attention} and feed-forward network (FFN) layers whose inputs are normalized~\cite{ba2016layer}.\footnote{All the illustrations in this paper are for the so-called \emph{pre-LayerNorm} configuration, in which the \emph{input} of the layer is normalized. Likewise, if the \emph{output} is normalized instead, the configuration is called \emph{post-LayerNorm}. Since pre-LayerNorm is known to perform better than post-LayerNorm~\cite{wang2019learning,nguyen2019transformers}, we only conducted experiments for the former, although our implementation supports both.} This is almost the same as the encoder of the original Transformer~\cite{vaswani2017attention} (we refer to the corresponding paper for further details), except that the embedding layer in our encoder is a small convolutional neural network (CNN)~\cite{fukushima1982neocognitron,lecun1989backpropagation} of two layers with ReLU activations and a stride of $2$, therefore reducing the input length by $4$.

The parallel dual-decoder consists of: (a) two decoders that follow closely the common Transformer decoder structure, and (b) four additional multi-head attention layers (called \emph{dual-attention} layers). Each dual-attention layer is complementary to a corresponding main attention layer. We recall that an attention layer receives as inputs a query $\Q\in\RR^{d_k}$, a key $\K\in\RR^{d_k}$, a value $\V\in\RR^{d_v}$ and outputs $\attention(\Q, \K, \V)$.\footnote{Following~\newcite{vaswani2017attention}, we use \emph{multi-head attention}, in which the inputs are linearly projected multiple times before feeding to the function, then the outputs are concatenated and projected back to the original dimension.}
A dual-attention layer receives $\Q$ from the main branch and $\K,\V$ from the other decoder at the same level (\ie at the same depth in Transformer architecture) %
to compute hidden representations that will be merged back into the main branch (\ie one decoder attends to the other in parallel). We present in more detail this merging operation in Section~\ref{sec:dd-variants}.

Our second proposed architecture is called \emph{cross dual-decoder Transformer}, which is similar to the previous one, except that now the dual-attention layers receive $\K,\V$ from the previous decoding step outputs of the other decoder, as illustrated in Figure~\ref{fig:dd-cross-semantic}. Thanks to this design, each prediction step can be performed separately on the two decoders. The hidden representations $\h^y_s, \h^z_t$ in~\eqref{eq:dd-embedding} produced by the decoders can thus be decomposed into:\footnote{This decomposition is clearly not possible for the parallel dual-decoder Transformer.}
\begin{align}
\h^y_s &= \decoder_\mr{asr}(\y_{< s},\z_{< t},\encoder(\x)) \in\RR^{d_y},\\
\h^z_t &= \decoder_\mr{st}(\z_{< t},\y_{< s},\encoder(\x)) \in\RR^{d_z}.
\end{align}

\subsection{Special cases}
\label{sec:degenerate-cases}

In this section, we present some special cases of our dual-decoder architecture and discuss their links to existing models in the literature.

\paragraph{Independent decoders}
When there is no dual-attention, the two decoders become \emph{independent}. In this case, the prediction joint probability can be factorized simply as
$p(y_s,z_t \mid \y_{< s},\z_{< t},\x)  = p(y_s \mid \y_{< s},\x)p(z_t \mid \z_{< t},\x).$
Therefore, \emph{all} prediction steps are separable and thus this model is the most computationally efficient. %
In the literature, this model is often referred to as \emph{multi-task} \cite{anastasopoulos2018tied,sperber2020consistent}.

\paragraph{Chained decoders}
Another special case corresponds to the extreme wait-$k$ policy, in which one decoder waits for the other to completely finish before starting its own decoding. For example, if ST waits for ASR, then the prediction joint probability reads $p(y_s,z_t \mid \y_{< s},\z_{< t},\x)  = p(y_s \mid \y_{< s},\x)p(z_t \mid \z_{< t},\y,\x).$ 
This model is called \emph{triangle} in previous work~\cite{anastasopoulos2018tied,sperber2020consistent}. A special case of this model is when the second decoder in the chain is not directly connected to the encoder, also referred to as \emph{two-stage}\footnote{
Also called \emph{cascade} by~\newcite{anastasopoulos2018tied}. We omit this term to avoid confusion with the common cascade models that are typically not trained end-to-end. Note that our chained-decoder (both \emph{triangle} and \emph{two-stage}) are end-to-end.}~\cite{sperber2019attention,sperber2020consistent}.

To summarize the different cases, we show
below the joint probability distributions encoded by the presented models, in decreasing level of dependencies: 
\begin{align}
&\text{(single output)} & p(\y,\z \mid \x)  &= \prod_{t=0}^{T} p(y_t,z_t \mid \y_{< t}, \z_{< t}, \x),\\
&\text{(dual-decoder)} & p(\y,\z \mid \x)  &= \prod_{t=0}^{T} p(y_t \mid \y_{< t}, \z_{< t}, \x) p(z_t \mid \y_{< t}, \z_{< t}, \x),\\
&\text{(chained)} & p(\y,\z \mid \x)  &= \prod_{t=0}^{T} p(y_t \mid \y_{< t}, \x) p(z_t \mid \z_{< t}, \y, \x),\\
&\text{(independent)} & p(\y,\z \mid \x)  &= \prod_{t=0}^{T} p(y_t \mid \y_{< t}, \x) p(z_t \mid \z_{< t}, \x),
\end{align}
where $T = \max(T_y,T_z)$. Similar formalization for the wait-$k$ policy~\eqref{eq:wait-k-distribution} can be obtained in a straightforward manner. Note that for independent decoders, the distribution is the same as in non-wait-$k$.

\subsection{Variants}\label{sec:dd-variants}
The previous section presents special cases of our formulation at a high level. In this section, we introduce different fine-grained variants of the dual-decoder Transformers used in the experiments (Section~\ref{sec:experiments}). %

\paragraph{Asymmetric dual-decoder} Instead of using all the dual-attention layers, one may want to allow a one-way attention: either ASR attends ST or the inverse, but not both.

\paragraph{At-self or at-source dual-attention} In each decoder block, 
there are two different attention layers, which we respectively call \emph{self-attention} (bottom) and \emph{source-attention}\footnote{{Also referred to as \emph{cross-attention} in the literature. We use a different name to avoid confusion with the \emph{cross} dual-decoder.}} (top). For each, there is an associated dual-attention, named respectively \emph{dual-attention at self} and \emph{dual-attention at source}. In the experiments, 
we study the case where either only the at-self or at-source attention layers are retained.

\paragraph{Merging operators}
The \var{Merge} layers shown in Figure~\ref{fig:dd-transformer} combine the outputs of the main attention $\H_{\mathrm{main}}$ and the dual-attention $\H_{\textrm{dual}}$. We experimented dual-attention with two different merging operators: weighted sum or concatenation. %
We can formally define the merging operators as
\begin{equation}\label{eq:merge}
    \H_{\mathrm{out}} = \merge(\H_{\mathrm{main}}, \H_{\mathrm{dual}}) \triangleq 
    \begin{cases}
    \H_{\mathrm{main}} &\mbox{if no dual-attention,}\\
    \H_{\mathrm{main}} + \lambda \H_{\mathrm{dual}}, &\mbox{if \var{sum} operator},\\
    \linear\left(\left[\H_{\mathrm{main}}; \H_{\mathrm{dual}}\right]\right) &\mbox{if \var{concat} operator.}
    \end{cases}
\end{equation}
For the \var{sum} operator, in particular, we perform experiments for \emph{learnable} or \emph{fixed} $\lambda$.

\begin{remark}
The model proposed by~\newcite{liu2019synchronous} is a special case of our \emph{cross} dual-decoder Transformer with no dual-attention at source, no layer normalization for the input embeddings (Figure~\ref{fig:dd-cross-semantic}), and \var{sum} merging with fixed $\lambda$.

\end{remark}

\section{Training and Decoding}

\subsection{Training}
\label{sec:training-methods}

The objective, $L(\hat{\y},\hat{\z},\y,\z) = \alpha L_{\mathrm{asr}}(\hat{\y},\y) + (1 - \alpha) L_{\mathrm{st}}(\hat{\z},\z)$, is a weighted sum of the cross-entropy ASR and ST losses,
where $(\hat{\y}, \hat{\z})$ and $(\y, \z)$ denote the predictions and the ground-truths for (ASR, ST), respectively. The weight $\alpha$ is set to $0.3$ in all experiments. Here we favor the ST task based on the intuition that it is more difficult to train than the ASR one, simply because of the multilinguality. A hyperparameter search may further improve the results. We also employ label smoothing~\cite{szegedy2016rethinking} with $\epsilon = 0.1$. For each language pair, training data is sorted by the number of frames. Each mini-batch contains all languages such that their numbers of frames are roughly the same. We follow \newcite{inaguma2019multilingual} and prepend a language-specific token to the target sentence. Preliminary experiments showed that this approach was more effective than adding a target language embedding along the temporal dimension to the speech feature inputs~\cite{di2019must}.

\subsection{Decoding}
\label{sec:decoding}
We present the beam search strategy used by our model.
Since there are two different outputs (ASR and ST), one may naturally think about two different beams (with possibly some interactions). However, we found that a \emph{single joint beam} works best for our model.
In this beam search strategy, each hypothesis includes a tuple of ASR and ST sub-hypotheses. The two sub-hypotheses are expanded together and the score is computed based on the sum of log probabilities of the output token pairs. For a beam size $B$, the $B$ best hypotheses are retained based on this score. In this setup, both sub-hypotheses evolve jointly,
which resembles the training process more than in the case of two different beams.
A limitation of this joint-beam strategy is that, in extreme cases,
one of the task (ASR or ST) may only have a single hypothesis. Indeed, at a decoding step $t+1$, we take the best $B$ predictions $(\hat{y}_{t},\hat{z}_{t})$
in terms of their sum of scores $s(y_t,z_t) \triangleq \log p(y_t\mid \y_{< t}, \z_{< t}) + \log p(z_t\mid \y_{< t}, \z_{< t})$; it can happen that, \eg, some $\hat{y}_t$ has a so dominant score that it is selected for all the hypotheses, \ie the $B$ (different) hypotheses have a single $\hat{y}_t$ and $B$ different $\hat{z}_t$. %
We leave the design of a joint-beam strategy with enforced diversity to future work. Finally, to produce translations for multiple target languages in our system, it suffices to feed different language-specific tokens to the dual-decoder at decoding time.

\section{Experiments}\label{sec:experiments}

\subsection{Dataset}
To build a one-to-many model that can jointly transcribe and translate, we use MuST-C~\cite{di2019must}, which is currently the largest publicly available one-to-many speech translation dataset.\footnote{Smaller datasets include Europarl-ST~\cite{iranzo2020europarl} and MaSS~\cite{boito2020mass}. Recently, a very large many-to-many dataset called CoVoST-2~\cite{wang2020covost2} has been released, while its predecessor CoVoST~\cite{wang2020covost} only covers the many-to-one scenario.}
MuST-C covers language pairs from English to eight different target languages including Dutch, French, German, Italian, Portuguese, Romanian, Russian, and Spanish. Each language direction includes a triplet of source input speech, source transcription, and target translation, with size ranging from 385 hours (Portuguese) to 504 hours (Spanish). We refer to the  original paper for more details.

\subsection{Training and decoding details}\label{sec:training-details}
Our implementation is based on the ESPnet-ST toolkit~\cite{inaguma2020espnet}.\footnote{\url{https://github.com/espnet/espnet}} %
In the following, we provide details for reproducing the results. The pipeline is identical for all experiments.

\paragraph{Models}
All experiments use the same encoder architecture with 12 layers.
The decoder has 6 layers, except for the independent-decoder model where we also include a 8-layer version (\var{independent$++$}) to compare the effects of dual-attention against simply increasing the number of model parameters.

\paragraph{Text pre-processing} Transcriptions and translations were normalized and tokenized using the Moses tokenizer~\cite{koehn2007moses}. The transcription was lower-cased and the punctuation was stripped. %
A joint BPE~\cite{sennrich2016neural} with 8000 merge operations was learned on the concatenation of the English transcription and all target languages. {We also experimented with two separate dictionaries (one for English and another for all target languages), but found that the results are worse.}

\paragraph{Speech features} We used Kaldi~\cite{povey2011kaldi} to extract 83-dimensional features (80-channel log Mel filter-bank coefficients and 3-dimensional pitch features) that were normalized by the mean and standard deviation computed on the training set.
Following common practice~\cite{inaguma2019multilingual,wang2020curriculum}, utterances having more than 3000 frames or more than 400 characters were removed. For data augmentation, we used \emph{speed pertubation}~\cite{ko2015audio} with three factors of $0.9$, $1.0$, and $1.1$ and \emph{SpecAugment}~\cite{daniel2019spec} with three types of deterioration including time warping ($W$), time masking ($T$) and frequency masking ($F$), where $W=5, T=40$, and $F=30$.

\paragraph{Optimization}
Following standard practice for training Transformer, we used the Adam optimizer~\cite{kingma2015adam} with Noam learning rate schedule~\cite{vaswani2017attention}, in which the learning rate is linearly increased for the first 25K warm-up steps then decreased proportionally to the inverse square root of the step counter. We set the initial learning rate to $\num{1e-3}$ and the Adam parameters to $\beta_1=0.9, \beta_2=0.98, \epsilon=\num{1e-9}$. We used a batch size of 32 sentences per GPU, with gradient accumulation of 2 training steps. All models were trained on a single-machine with 8 32GB GPUs for 250K steps unless otherwise specified. As for model initialization, we trained an independent-decoder model with the two decoders having \emph{shared} weights for 150K steps and used its weights to initialize the other models. This resulted in much faster convergence for all models. We also included this shared model in the experiments, and for a fair comparison, we trained it for additional 250K steps. Finally, for decoding, we used a beam size of $10$ with length penalty of $0.5$.\footnote{For a hypothesis of length $L$, a length penalty $p$ means a score of $pL$ will be added to the (log probability) score of that hypothesis (Section~\ref{sec:decoding}). Therefore, longer hypotheses are favored, or equivalently, shorter hypotheses are ``penalized''.}

\subsection{Results and analysis}

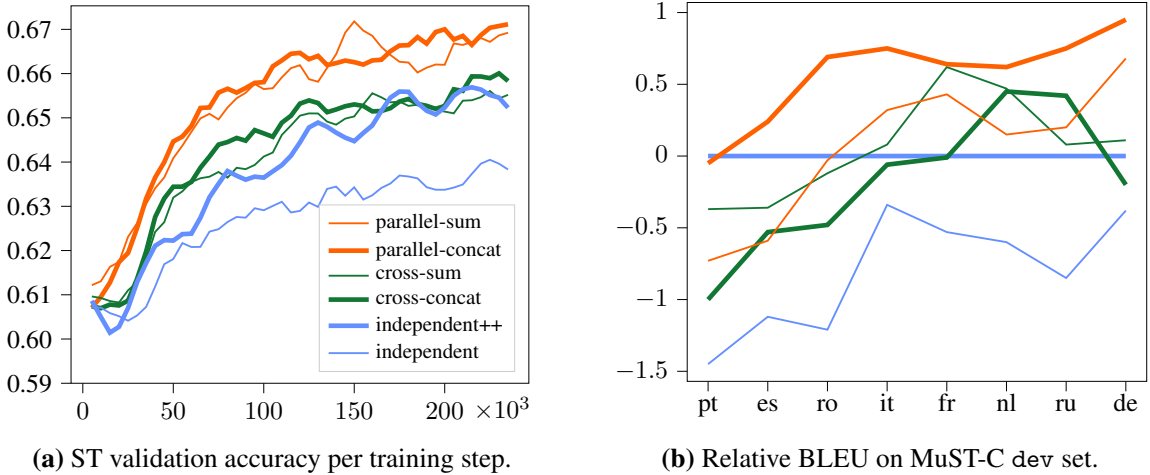
\begin{figure}[!tb]
\centering
\begin{subfigure}[b]{.5\textwidth}
    \centering
    \resizebox{0.9\textwidth}{!}{\input{validation_acc}}
    \caption{ST validation accuracy per training step.}
    \label{fig:validation-acc}
\end{subfigure}%
\begin{subfigure}[b]{.5\textwidth}
  \centering
  \resizebox{0.9\textwidth}{!}{\input{dev_bleu}}
  \caption{Relative BLEU on MuST-C \var{dev} set.}
  \label{fig:bleu-dev}
\end{subfigure}
\caption{Results on the MuST-C \var{dev} set. The models listed in the legends correspond respectively to lines 17, 16, 8, 7, 3, 2 in Table~\ref{tbl:bleu_dev_bs}. The baseline used for relative BLEU is \var{independent++}. One can observe that the parallel models consistently outperform the others in terms of validation accuracy.
}
\label{fig:dd-vs-independent}
\end{figure}

In this section, we report detokenized case-sensitive BLEU\footnote{{We also tried sacreBLEU~\cite{post2018call} and found that the results are identical.}}~\cite{papineni2002bleu} on the MuST-C \var{dev} sets (Table~\ref{tbl:bleu_dev_bs}). Results on the \var{test} sets are discussed in Section~\ref{section:sota}. {Following previous work~\cite{inaguma2020espnet}, we remove non-verbal tokens in evaluation.\footnote{This is for a fair comparison with the results of \newcite{inaguma2020espnet}, presented in Section~\ref{section:sota}.}} In Table~\ref{tbl:bleu_dev_bs}, there are 3 main groups of models, corresponding to independent-decoder, cross dual-decoder (\var{crx}), and parallel dual-decoder (\var{par}), respectively. In particular, \var{independent++} corresponds to a 8-decoder-layer model and will serve as our strongest baseline for comparison. Figure~\ref{fig:dd-vs-independent} shows the relative performance of some representative models with respect to this baseline, together with their validation accuracies. In the following, when comparing models, we implicitly mean ``on average'' (over the 8 languages), except otherwise specified.

\begin{table}[!htb]
    \centering
    \resizebox{\textwidth}{!}{
    \begin{tabular}{l | lll ll | l | lll lll ll | l | l}
        \toprule
         \textbf{No} & \textbf{type} & \textbf{side} & \textbf{self} & \textbf{src} & \textbf{merge} & \textbf{params} &
        \textbf{de} & \textbf{es} & \textbf{fr} & \textbf{it} & \textbf{nl} & \textbf{pt} & \textbf{ro} & \textbf{ru} &
        \textbf{avg} & \textbf{WER}\\
        \midrule
        
        1 & \multicolumn{4}{l}{independent (shared)} & & 31.3M & 19.40 & 27.77 & 24.65 & 19.93 & 21.53 & 24.24 & 18.19 & 10.99 & 20.84 & 14.2 \\
        2 & \multicolumn{4}{l}{independent} & & 44.8M & 20.11 & 28.18 & 25.61 & 20.76 & 21.83 & 25.45 & 18.45 & 11.31 & 21.46 & 12.6 \\
        3 & \multicolumn{4}{l}{independent++} & & 51.2M & 20.25 & 29.48 & 26.10 & 21.05 & 22.34 & \textbf{26.71} & 19.67 & 12.10 & 22.21 & 12.9\\
        \midrule
        
       4 & crx & st & - & \checkmark & sum & 46.4M & 20.01 & 28.57 & 25.86 & 20.66 &  {22.26} & 25.36 & 19.06 & 12.00 & 21.72 & \highlight{12.7}\\
       5 & crx & both & - & \checkmark & concat & 51.2M &  \highlight{20.36} & 28.51 & 25.80 &  \highlight{21.18} & 22.10 & 25.24 & 19.55 & 11.89 & 21.83 & \highlight{12.3}\\
       6 & crx & both & - & \checkmark & sum & 48.0M & 19.99 & 28.87 &  \highlight{26.09} & 20.94 & 21.67 & 25.42 & 18.85 & 11.83 & 21.71 & \highlight{12.2}\\
       7 & crx & both & \checkmark & \checkmark & concat & 54.3M & 20.07 & 28.73 & 26.01 & 20.93 &  \highlight{22.59} & 25.60 & 19.08 &  \highlight{12.46} & 21.93 & \highlight{12.4}\\
       8 & crx & both & \checkmark & \checkmark & sum & 51.2M &  \highlight{20.38} & 28.90 &  \highlight{26.64} &  \highlight{21.07} &  \highlight{22.61} & 26.23 & 19.44 &  \highlight{12.12} &  {22.17} & \highlight{\textbf{12.1}}\\
       9 & crx\textsuperscript{$\star$} & both & \checkmark & - & sum & 48.0M & 19.72 & 27.96 & 25.49 & 20.52 & 21.56 & 25.01 & 18.53 & 11.33 & 21.26 & \highlight{12.8}\\
       10 & crx\textsuperscript{$\star$} & both & \checkmark & - & sum\textsuperscript{$\dagger$} & 48.0M & 18.62 & 27.11 & 24.41 & 19.73 & 20.47 & 24.49 & 17.23 & 11.09 & 20.39 & \highlight{12.8}\\
       11 & crx\textsuperscript{$\star$} & both & \checkmark & \checkmark & sum & 51.2M & 19.54 & 28.17 & 25.68 & 20.95 & 21.55 & 24.77 & 18.76 & 11.28 & 21.34 & \highlight{12.3}\\
       \midrule
       12 & par & st & \checkmark & \checkmark & concat & 49.6M &  \highlight{20.57} & 28.84 &  {26.08} & 20.85 & 22.11 & 25.70 & 19.36 & 11.90 & 21.93 & 13.0\\
       13 & par & both & - & \checkmark & concat & 51.2M &  \highlight{20.84} & \highlight{29.51} &  \highlight{26.44} &  \highlight{21.53} &  \highlight{22.68} & 25.94 & 19.04 &  \highlight{12.60} &  \highlight{22.32} & \highlight{12.5}\\
       14 & par & both & - & \checkmark & sum & 48.0M &  \highlight{20.85} & 29.18 &  \highlight{26.38} & \highlight{\textbf{22.14}} & \highlight{22.87} & 26.49 &  \highlight{19.70} &  \highlight{12.74} &  \highlight{22.54} & \highlight{12.7}\\
       15 & par & both & \checkmark & - & sum & 48.0M &  \highlight{20.56} & 29.21 &  \highlight{26.54} &  \highlight{21.07} &  \highlight{22.51} & 25.75 &  {19.64} & \highlight{\textbf{12.80}} &  \highlight{22.26} & \highlight{12.8}\\
       16 & par & both & \checkmark & \checkmark & concat & 54.3M & \highlight{\textbf{21.22}} &  \highlight{29.50} & \highlight{\textbf{26.66}} &  \highlight{21.74} &  \highlight{22.76} & {26.66} & \highlight{\textbf{20.25}} &  \highlight{12.79} & \highlight{22.70} & \highlight{12.7}\\
       17 & par & both & \checkmark & \checkmark & sum & 51.2M &  \highlight{20.95} & 28.67 &  \highlight{26.45} &  \highlight{21.31} &  \highlight{22.29} & 25.87 & 19.53 &  \highlight{12.24} &  {22.16} & \highlight{12.8}\\
       18 & par & both\textsuperscript{R3} & - & \checkmark & sum & 48.0M & \highlight{\textbf{21.22}} & \highlight{\textbf{30.12}} & \highlight{26.53} & \highlight{22.06} & \highlight{\textbf{23.37}} & 26.59 & \highlight{19.82} & \highlight{12.54} & \highlight{\textbf{22.78}} & \highlight{12.6}\\
        19 & par & both\textsuperscript{T3} & - & \checkmark & sum & 48.0M & 20.35 & 28.61 & 25.94 & 21.22 & 22.12 & 25.19 & 19.36 & 11.99 & 21.85 & 13.6\\
       
    \bottomrule
    \end{tabular}
    }
    {\footnotesize
    \begin{flushleft}
	$^\star$no normalization for dual-attention input, $^\dagger$\var{sum} merging has $\lambda = 0.3$ fixed,\\
	{$^\text{R3}$ASR is 3 steps ahead of ST, $^\text{T3}$ST is 3 steps ahead of ASR}.
    \end{flushleft}
    
    }
    
    \caption{BLEU and (average) WER on MuST-C dev set. In the second column (\var{type}), \var{crx} and \var{parallel} denote the cross and parallel dual-decoder, respectively. In the third column (\var{side}), \var{st} means only ST attends to ASR. Line 1 corresponds to the independent-decoder model where the weights of the two decoders are shared, and \var{independent++} corresponds to the model with 8 decoder layers (instead of 6). It should be noted that line 10 corresponds to the model proposed by~\protect\newcite{liu2019synchronous}. The values that are better than the baseline (\var{independent++}) are {underlined} and colored in blue, while the best values are highlighted in bold.
    }
    \label{tbl:bleu_dev_bs} \vspace*{-3pt}
\end{table}

\paragraph{Parallel \vs cross} Under the same configurations, parallel models outperform their cross counterparts in terms of translation (line 5 \vs line 13, line 6 \vs line 14, and line 7 \vs line 16), showing an improvement of $0.7$ BLEU on average. In terms of recognition, however, the parallel architecture has on average a 0.33\% higher (worse) WER compared to the cross models. %
On the other hand, parallel dual-decoders perform better than independent decoders in both translation and recognition tasks, except for the asymmetric case (line 12), the at-self and at-source with \var{sum} merging configuration (line 17), and the wait-$k$ model where ST is ahead of ASR (line 19). This shows that both the translation and recognition tasks can benefit from the tight interaction between the two decoders, \ie it is possible to achieve no trade-off between BLEUs and WERs for the parallel models compared to the independent architecture. This is not the case, however, for the cross dual-decoders that feature weaker interaction than the parallel ones. Interestingly, there is a slight trade-off between the parallel and cross designs: the parallel models are better in terms of BLEUs but worse in terms of WERs. This is to some extent similar to previous work where models having different types of trade-offs between BLEUs and WERs~\cite{he2011word,sperber2020consistent,chuang2020worse}. It should be emphasized that most of the dual-decoder models have fewer or same numbers of parameters compared to \var{independent++}. This confirms our intuition that the tight connection between the two decoders in the parallel architecture improves performance. %
The cross dual-decoders perform relatively well compared to the baseline of two independent decoders with the same number of layers (6), but not so well compared to the 
stronger baseline with 8 layers.%

\paragraph{Symmetric \vs asymmetric}
In some experiments, we only allow the ST decoder to attend to the ASR decoder. For the cross dual-decoder, this did not yield {noticeable} improvements in terms of BLEU ($21.72$ at line 4 \vs $21.71$ at line 6), while for the parallel architecture, the results are worse ($21.93$ at line 12 \vs $22.70$ at line 16). {The symmetric models also outperform the asymmetric counterparts in terms of WER ($12.7$ at line 4 \vs $12.2$ at line 6, $13.0$ at line 12 \vs $12.7$ at line 16). It is confirmed again that the two tasks are complementary and can help each other: removing the ASR-to-ST attention hurts performance. In fact, examining the learnable $\lambda$ in the \var{sum} merging operator shows that the decoders learn to attend to each other, though at different rates. We observed that for the same layer depth, the ST decoder always attends more to the ASR one, and for both of them $\lambda$ increases with the depth of the layer.}%

\paragraph{At-self dual-attention \vs at-source dual-attention}
For the parallel dual-decoder, the at-source dual-attention produces better results than the at-self counterpart (BLEU: $22.54$ \vs $22.26$, WER: $12.7$ \vs $12.8$ at line 14 \vs line 15), while the combination of both does not improve the results (BLEU $22.16$, WER $12.8$ at line 17). For the \var{concat} merging, using both yields better results in terms of translation but slightly hurts the recognition task (BLEU: $22.70$ \vs $22.32$, WER: $12.7$ \vs $12.5$ at line 16 \vs line 13). %

\paragraph{Sum \vs concat merging}
The impact of merging operators is not consistent across different models. 
If we focus on the parallel dual-decoder, \var{sum} is better for models with only at-source attention (line 13 \vs line 14) and \var{concat} is better for models using both at-self and at-source attention (line 16 \vs line 17). 

\paragraph{Input normalization and learnable sum} 
Some experiments confirm the importance of normalizing the input fed to the dual-attention layers (\ie the LayerNorm layers shown in Figure~\ref{fig:dd-cross-semantic}).
The results show that normalization {substantially} improves the performance (BLEU: $22.17$ \vs $21.34$, WER: $12.1$ \vs $12.3$ at line 8 \vs at line 11). It is also beneficial to use learnable weights compared to a fixed value for the \var{sum} merging operator (Equation~\eqref{eq:merge}) (BLEU: $21.26$ \vs $20.39$ at line 9 \vs line 10). Note that the fixed weight and non-normalization configuration corresponds to the model of \newcite{liu2019synchronous} (line 10).

\paragraph{Wait-$k$ policy} %
We compare a non-wait-$k$ parallel dual-decoder (line 14) with its wait-$k$ ($k=3$) counterparts. From the results, one can observe that letting ASR be ahead of ST (line 18) improves the performance ({BLEU: }$22.78$ \vs $22.54${, WER: $12.6$ \vs $12.7$}), while letting ST be ahead of ASR (line 19) {considerably} worsen the results ({BLEU: }$21.85${, WER: $13.6$}). This confirms our intuition that the ST task is more difficult and should not take the lead in the dual-decoder models. %

\paragraph{ASR results} From the results (last column of Table~\ref{tbl:bleu_dev_bs}), one can observe that the dual-decoder models outperform the baseline \var{indepedent++}, except for the asymmetric case and the wait-$k$ model where ST is 3 steps ahead of ASR. While using a single decoder leads to an average of 14.2\% WER, all other symmetric architectures with two decoders (except the ASR-waits-for-ST) have better and rather stable WERs (from 12.1\% to 13.0\%). Detailed results for each data subset are provided in the Appendix.

\subsection{Comparison to state-of-the-art}
\label{section:sota}

To avoid a hyper-parameter search over the test set, we only select three of our best models together with the baseline \var{independent++} for evaluation. All of the three models are symmetric parallel dual-decoders, the first one has at-source dual-attention with \var{sum} merging, the second one has both at-self and at-source dual-attentions with \var{concat} merging, and the last one is a wait-$k$ model in which ASR is 3 steps ahead of ST. These models correspond to lines 5, 6, and 7 of Table~\ref{tbl:bleu_test_bs}, and will be referred to respectively as \var{par++}, \var{par}, and \var{par$^{\textrm{R3}}$} in the sequel. For \var{par++} we increase the number of decoder layers from 6 to 8, thus increasing the number of parameters from 48M to 51.2M, matching that of the baseline. We do not do this for \var{par$^{\textrm{R3}}$} (48M) as this model already has a higher latency due to the wait-$k$. 
All models are trained for 550K steps, corresponding to 25 epochs. 
Following~\newcite{inaguma2020espnet}, we use the average of five checkpoints with the best validation accuracies on the dev sets for evaluation.

\begin{table}[!htb]
    \centering
    \resizebox{\textwidth}{!}{
    \begin{tabular}{l | lll ll  c | lll lll ll | l| l}
        \toprule
         \textbf{No} & \textbf{type} & \textbf{side} & \textbf{self} & \textbf{src} & \textbf{merge} & \textbf{epochs} &
        \textbf{de} & \textbf{es} & \textbf{fr} & \textbf{it} & \textbf{nl} & \textbf{pt} & \textbf{ro} & \textbf{ru} &
        \textbf{avg} & \textbf{WER}\\
        \midrule
        
        1 & \multicolumn{5}{l}{Bilingual one-to-one~\cite{inaguma2020espnet}} & 50 & 22.91 & 27.96 & 32.69 & 23.75 & 27.43 & 28.01 & 21.90 & \textbf{15.75} & 25.05 & 12.0\\
        \midrule
        2 & \multicolumn{5}{l}{Multilingual one-to-many~\cite{di2019one}} &  & 17.70 & 20.90 & 26.50 & 18.00 & 20.00 & 22.60 & - & - & - & -\\
        3 & \multicolumn{5}{l}{Multilingual one-to-many~\cite{di2019one}} &  & 16.50 & 18.90 & 24.50 & 16.20 & 17.80 & 20.80 & 15.90 & 9.80 & 17.55 & -\\
        \midrule
        
        4 & \multicolumn{5}{l}{independent++} & 25 & 22.82 &27.20 &32.11 &23.34 &26.67 &28.98 &21.37 &14.34 &24.60 & 11.6\\
        \midrule
        
        5 & par++ & both & - & \checkmark & sum & 25 & \textbf{23.63} &\textbf{28.12} &\textbf{33.45} &\textbf{24.18} &\textbf{27.55} &\textbf{29.95} &\textbf{22.87} &15.21 &\textbf{25.62} & \textbf{11.4} \\
        
        6 & par & both & \checkmark & \checkmark & concat & 25 & 22.74 &27.59 &32.86 &23.50 &26.97 &29.51 &21.94 &14.88 &25.00 & 11.6\\
         
        7 & par$^{\textrm{R3}}$ & both & - & \checkmark & sum & 25 & 22.84 &27.92 &32.12 &23.61 &27.29 &29.48 &21.16 &14.50 &24.87 & 11.6\\
        
    \bottomrule
    \end{tabular}
        
    }
    \caption[BLEU on MuST-C \var{tst-COMMON} test set.]{BLEU on MuST-C \var{tst-COMMON} test set. Line 2 corresponds to the best multilingual models of~\protect\newcite{di2019one} trained separately for \var{\{de,nl\}} and \var{\{es,fr,it,pt\}}, while line 3 is a single model trained on 8 languages. %
    The WER in line 1 corresponds to the best result reported by~\newcite{inaguma2020espnet}.\footnotemark
    }
    \label{tbl:bleu_test_bs}
\end{table}

We compare the results with the previous work~\cite{di2019one} in the multilingual setting. In addition, to demonstrate the competitive performance of our models, we also include the best existing translation performance on MuST-C~\cite{inaguma2020espnet}, although these results were obtained with bilingual systems and from a sophisticated training recipe. Indeed, to obtain the results for each language pair (\eg \var{en-de}), \newcite{inaguma2020espnet} pre-trained an ASR model and an MT model to initialize the weights of (respectively) the encoder and decoder for ST training. This means that to obtain the results for the 8 language pairs, 24 independent trainings had to be performed in total (3 for each language pair).

The results in Table~\ref{tbl:bleu_test_bs} show that our models achieved very competitive performance compared to bilingual one-to-one models \cite{inaguma2020espnet}, despite being trained for only half the number of epochs. In particular, the \var{par++} model achieved the best results, consistently surpassing the others on all languages (except on Russian where it is outperformed by the bilingual model). Our results also surpassed those of \newcite{di2019one} by a large margin. We observe the largest improvements on Portuguese ($+1.94$ at line 5, $+1.50$ at line 6, and $+1.47$ at line 7, compared to the bilingual result at line 1), which has the least data among the 8 language pairs in MuST-C. This phenomenon is also common in multilingual neural machine translation where multilingual joint training has been shown to improve performance on low-resource languages~\cite{johnson2017google}. %
\footnotetext{\url{https://github.com/espnet/espnet/blob/master/egs/must_c/asr1/RESULTS.md}}

\section{Conclusion}

We introduced a novel dual-decoder Transformer architecture for synchronous speech recognition and multilingual speech translation. Through a dual-attention mechanism, the decoders in this model are at the same time able to specialize in their tasks while being helpful to each other. 
The proposed model also generalizes previously proposed approaches using two independent (or {weakly tied}) decoders or chaining ASR and ST. It is also flexible enough to experiment with settings where ASR is ahead of ST which makes it promising for (one-to-many) simultaneous speech translation.
Experiments on the MuST-C dataset showed that our model achieved very competitive performance compared to state-of-the-art.

\section*{Acknowledgements}
{This work was supported by a Facebook AI SRA grant, and was granted access to the HPC resources of IDRIS under the allocations 2020-AD011011695 and 2020-AP011011765 made by GENCI. It was also done as part of the Multidisciplinary Institute in Artificial Intelligence MIAI@Grenoble-Alpes (ANR-19-P3IA-0003). We thank the anonymous reviewers for their insightful feedback.}

\bibliographystyle{coling}
\bibliography{references,mybib,refs}

\section*{Appendix}

\begin{figure}[!htb]
    \centering
    \includegraphics[width=0.8\textwidth]{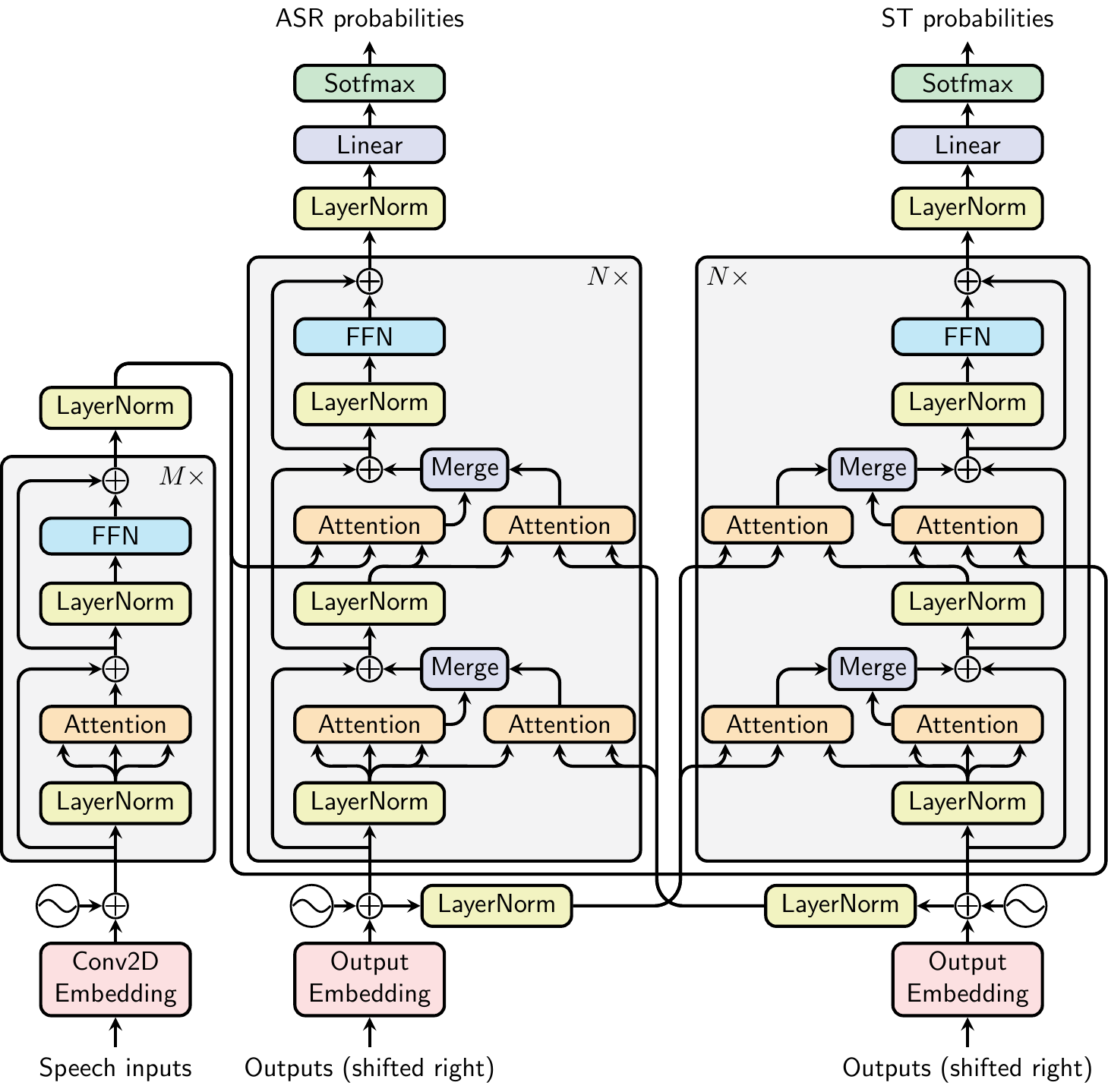}
    \caption{The \emph{cross} dual-decoder Transformer. Unlike the \emph{parallel} dual-decoder Transformer, here one decoder attends to the previous decoding-step outputs of the other and there is no interaction between their hidden states.}
    \label{fig:dd_cross}
\end{figure}

We present the detailed model architecture for the \emph{cross dual-decoder Transformer} in Figure~\ref{fig:dd_cross}. Detailed WER results on MuST-C dev set are presented in Table~\ref{tbl:wer_dev_bs}. %

\begin{table}[!htb]
    \centering
    \resizebox{\textwidth}{!}{
    
\begin{tabular}{l | lll ll | l | lll lll ll | l }
	\toprule
	\textbf{No} & \textbf{type} & \textbf{side} & \textbf{self} & \textbf{src} & \textbf{merge} & \textbf{params} &
	\textbf{de} & \textbf{es} & \textbf{fr} & \textbf{it} & \textbf{nl} & \textbf{pt} & \textbf{ro} & \textbf{ru} &
	\textbf{avg} \\
	\midrule
	
    1 & \multicolumn{4}{l}{independent (shared)} & & 31.3M & 14.9 & 13.8 & 14.5 & 13.6 & 14.0 & 13.6 & 14.8 & 14.0 & 14.2\\
    2 & \multicolumn{4}{l}{independent} & & 44.8M & \highlight{12.6}& 12.6& \highlight{13.1}& \highlight{11.9}& \highlight{12.4}& \highlight{12.3}& \highlight{13.1}& \highlight{12.4}& \highlight{12.6}\\
    3 & \multicolumn{4}{l}{independent++} & & 51.2M & 13.4 & 12.5 & 14.0 & 12.2 & 12.7 & 12.8 & 13.3 & 12.5 & 12.9\\
	\midrule
	4 & crx & st & - & \checkmark & sum & 46.4M & \highlight{12.8}& 13.0& \highlight{13.2}& 12.8& \highlight{12.1}& \highlight{12.4}& \highlight{12.2}& 12.8& \highlight{12.7}\\
	5 & crx & both & - & \checkmark & concat & 51.2M & \highlight{12.3}& 12.9& \highlight{12.4}& \highlight{11.9}& \highlight{12.3}& \highlight{12.3}& \highlight{12.2}& \highlight{12.4}& \highlight{12.3}\\
	6 & crx & both & - & \checkmark & sum & 48.0M & \highlight{12.2}& 12.6& \highlight{12.4}& \highlight{12.0}& \highlight{12.3}& \highlight{12.0}& \highlight{12.0}& \highlight{12.2}& \highlight{12.2}\\
	7 & crx & both & \checkmark & \checkmark & concat & 54.3M & \highlight{12.5}& 12.9& \highlight{12.6}& \highlight{12.0}& \highlight{12.2}& \highlight{12.3}& \highlight{12.4}& \highlight{12.2}& \highlight{12.4}\\
	8 & crx & both & \checkmark & \checkmark & sum & 51.2M & \highlight{12.1}& 12.7& \highlight{12.1}& \highlight{11.7}& \highlight{12.0}& \highlight{12.0}& \highlight{12.0}& \highlight{12.1}& \highlight{12.1}\\
	9 & crx\textsuperscript{$\star$} & both & \checkmark & - & sum & 48.0M & \highlight{12.7}& 12.9& \highlight{13.3}& 12.5& \highlight{12.6}& \highlight{12.5}& \highlight{12.6}& 12.9& \highlight{12.8}\\
	10 & crx\textsuperscript{$\star$} & both & \checkmark & - & sum\textsuperscript{$\dagger$}  & 48.0M & \highlight{12.7}& 12.8& \highlight{13.2}& 12.3& \highlight{11.5}& 13.1& \highlight{12.7}& 12.9& \highlight{12.8}\\
	11 & crx\textsuperscript{$\star$} & both & \checkmark & \checkmark & sum& 51.2M & \highlight{12.2}& 12.6& \highlight{12.4}& \highlight{11.8}& \highlight{12.1}& \highlight{12.2}& \highlight{12.4}& \highlight{12.4}& \highlight{12.3}\\
	\midrule
	12 & par & st & \checkmark & \checkmark & concat & 49.6M & \highlight{13.0}& 13.7& \highlight{13.1}& 12.5& 13.1& \highlight{12.7}& 13.3& 12.9& 13.0\\
	13 & par & both & - & \checkmark & concat & 51.2M & \highlight{12.6}& 13.0& \highlight{12.6}& \highlight{12.1}& \highlight{12.4}& \highlight{12.3}& \highlight{12.5}& 12.5& \highlight{12.5}\\
	14 & par & both & - & \checkmark & sum & 48.0M & \highlight{12.8}& 13.0& \highlight{13.2}& 12.2& 12.8& \highlight{12.4}& \highlight{12.6}& 12.6& \highlight{12.7}\\
	15 & par & both\textsuperscript{R3} & - & \checkmark & sum & 48.0M & \highlight{13.1}& 13.4& \highlight{12.8}& 12.3& 12.9& \highlight{12.5}& \highlight{12.7}& 12.9& \highlight{12.8}\\
	16 & par & both\textsuperscript{T3} & - & \checkmark & sum & 48.0M & 14.0& 14.3& \highlight{13.9}& 13.4& 13.9& 13.6& 13.8& 13.5& 13.8\\
	17 & par & both & \checkmark & - & sum & 48.0M & \highlight{12.6}& 13.1& \highlight{13.3}& 12.4& \highlight{12.6}& \highlight{12.6}& \highlight{12.6}& 12.8& \highlight{12.8}\\
	18 & par & both & \checkmark & \checkmark & concat & 54.3M & \highlight{12.5}& 13.3& \highlight{12.9}& 12.3& 12.7& \highlight{12.6}& \highlight{12.6}& 12.5& \highlight{12.7}\\
	19 & par & both & \checkmark & \checkmark & sum & 51.2M &\highlight{12.7}& 13.4& \highlight{12.8}& 12.5& 12.8& \highlight{12.6}& \highlight{12.7}& 12.7& \highlight{12.8}\\
	\bottomrule
	\end{tabular}
    }
    \caption{Word error rate on MuST-C \texttt{dev} set. {The values that are better than the baseline (\var{independent++}) are underlined and colored in blue.}}
    \label{tbl:wer_dev_bs}
\end{table}

\end{document}

%% file: math.tex
\usepackage{amsmath}
\usepackage{amssymb}
\usepackage{bold-extra}
\usepackage{mdframed}
\usepackage{amsthm}
\usepackage{bbm}
\usepackage{bm}
\usepackage{accents} %

\theoremstyle{plain}

\newtheorem*{lemma*}{Lemma}

\newtheorem*{subproblem*}{Subproblem}

\newtheorem*{problem*}{Problem}

\newtheorem*{conjecture*}{Conjecture}

\theoremstyle{definition}

\newtheorem*{definition*}{Definition}

\newtheorem*{remark}{Remark}

\newcommand{\RR}{{\mathbb{R}}}

\ifdefined\b
\let\svb\b
\DeclareRobustCommand\b{\ifmmode\mathbf{b}\else\expandafter\svb\fi}
\else
\newcommand{\b}{{\mathbf{b}}}
\fi

\ifdefined\c
	\let\svc\c
	\DeclareRobustCommand\c{\ifmmode\mathbf{c}\else\expandafter\svc\fi}
\else
	\newcommand{\c}{{\mathbf{c}}}
\fi

\ifdefined\k
\let\svk\k
\DeclareRobustCommand\k{\ifmmode\mathbf{k}\else\expandafter\svk\fi}
\else
\newcommand{\k}{{\mathbf{k}}}
\fi

\ifdefined\v
	\let\svv\v
	\DeclareRobustCommand\v{\ifmmode\mathbf{v}\else\expandafter\svv\fi}
\else
	\newcommand{\v}{{\mathbf{v}}}
\fi

\ifdefined\u
	\let\svu\u
	\DeclareRobustCommand\u{\ifmmode\mathbf{u}\else\expandafter\svu\fi}
\else
	\newcommand{\u}{{\mathbf{u}}}
\fi

\ifdefined\m
	\let\svm\m
	\DeclareRobustCommand\m{\ifmmode\mathbf{m}\else\expandafter\svm\fi}
\else
	\newcommand{\m}{{\mathbf{m}}}
\fi

\ifdefined\o
\let\svo\o
\DeclareRobustCommand\o{\ifmmode\mathbf{o}\else\expandafter\svo\fi}
\else
\newcommand{\o}{{\mathbf{o}}}
\fi

\let\svq\q
\DeclareRobustCommand\q{\ifmmode\mathbf{q}\else\expandafter\svq\fi}

\newcommand{\h}{{\mathbf{h}}}

\newcommand{\x}{{\mathbf{x}}}
\newcommand{\y}{{\mathbf{y}}}
\newcommand{\z}{{\mathbf{z}}}

\ifdefined\B
\renewcommand{\B}{{\mathbf{B}}}
\else
\newcommand{\B}{{\mathbf{B}}}
\fi

\ifdefined\C
\renewcommand{\C}{{\mathbf{C}}}
\else
\newcommand{\C}{{\mathbf{C}}}
\fi

\ifdefined\G
\let\svG\G
\DeclareRobustCommand\G{\ifmmode\mathbf{G}\else\expandafter\svG\fi}
\else
\newcommand{\G}{{\mathbf{G}}}
\fi

\ifdefined\H
\let\svH\H
\DeclareRobustCommand\H{\ifmmode\mathbf{H}\else\expandafter\svH\fi}
\else
\newcommand{\H}{{\mathbf{H}}}
\fi

\ifdefined\M
\let\svH\M
\DeclareRobustCommand\M{\ifmmode\mathbf{M}\else\expandafter\svM\fi}
\else
\newcommand{\M}{{\mathbf{M}}}
\fi

\newcommand{\K}{{\mathbf{K}}}

\newcommand{\Q}{{\mathbf{Q}}}

\newcommand{\V}{{\mathbf{V}}}
\newcommand{\W}{{\mathbf{W}}}

\makeatletter
\newcommand*\dotp{\mathpalette\dotp@{.5}}
\newcommand*\dotp@[2]{\mathbin{\vcenter{\hbox{\scalebox{#2}{$\m@th#1\bullet$}}}}}
\makeatother

\newcommand{\mr}[1]{{\mathrm{#1}}}

\DeclareMathOperator{\attention}{Attention}

\DeclareMathOperator{\softmax}{softmax}
\DeclareMathOperator{\linear}{linear}
\DeclareMathOperator{\merge}{Merge}
\DeclareMathOperator{\encoder}{\textsc{Encoder}}
\DeclareMathOperator{\decoder}{\textsc{Decoder}}

%% file: validation_acc.tex
\begin{tikzpicture}

\definecolor{color0}{RGB}{254,97,0}
\definecolor{color1}{RGB}{254,97,0}
\definecolor{color2}{RGB}{17,119,51}
\definecolor{color3}{RGB}{17,119,51}
\definecolor{color4}{RGB}{100,143,255}
\definecolor{color5}{RGB}{100,143,255}

\begin{axis}[
legend cell align={left},
legend style={fill opacity=0.8, draw opacity=1, text opacity=1, at={(0.97,0.03)}, anchor=south east, draw=white!80!black, font=\footnotesize},
tick align=outside,
tick pos=left,
x grid style={white!69.0196078431373!black},
xmin=-6500, xmax=246500,
xtick style={color=black},
y grid style={white!69.0196078431373!black},
ymin=0.59, ymax=0.675316682457924,
ytick style={color=black},
ytick={0.59,0.6,0.61,0.62,0.63,0.64,0.65,0.66,0.67,0.68},
yticklabels={0.59,0.60,0.61,0.62,0.63,0.64,0.65,0.66,0.67,0.68},
scaled x ticks={real:1000},
xtick scale label code/.code={$\times 10^3$},
every x tick scale label/.style={at={(xticklabel cs:0.86)},yshift=-1.pt,anchor=south west}
]
\addplot [thick,color0]
table {%
5000 0.612174361944199
10000 0.613064050674438
15000 0.616322800517082
20000 0.617403790354729
25000 0.623099356889725
30000 0.626177474856377
35000 0.630540087819099
40000 0.634090512990952
45000 0.636511385440826
50000 0.640882030129433
55000 0.643688231706619
60000 0.646822884678841
65000 0.64987587928772
70000 0.650903776288033
75000 0.64963985979557
80000 0.652344733476639
85000 0.654428228735924
90000 0.655660599470139
95000 0.657644838094711
100000 0.656490311026573
105000 0.656703591346741
110000 0.659147053956985
115000 0.661224603652954
120000 0.661898747086525
125000 0.658739671111107
130000 0.658154636621475
135000 0.66128446161747
140000 0.664374142885208
145000 0.669239953160286
150000 0.671798840165138
155000 0.669722229242325
160000 0.66853292286396
165000 0.666022539138794
170000 0.664206251502037
175000 0.662674367427826
180000 0.662496238946915
185000 0.660239204764366
190000 0.661331564188004
195000 0.662051737308502
200000 0.66198767721653
205000 0.666826233267784
210000 0.666474938392639
215000 0.667129844427109
220000 0.668032541871071
225000 0.667168796062469
230000 0.668641000986099
235000 0.669233471155167
};
\addlegendentry{parallel-sum}
\addplot [line width=2.0pt, color1]
table {%
5000 0.607067659497261
10000 0.609555721282959
15000 0.612765744328499
20000 0.617386803030968
25000 0.619445770978928
30000 0.625085443258286
35000 0.63173596560955
40000 0.636526018381119
45000 0.639977246522903
50000 0.644658759236336
55000 0.645752802491188
60000 0.648137867450714
65000 0.652226775884628
70000 0.652339220046997
75000 0.655706271529198
80000 0.656573072075844
85000 0.655703440308571
90000 0.656740099191666
95000 0.657904550433159
100000 0.65806320309639
105000 0.661642238497734
110000 0.662977948784828
115000 0.664475157856941
120000 0.664688304066658
125000 0.663247436285019
130000 0.66400671005249
135000 0.66192652285099
140000 0.662370607256889
145000 0.662938311696053
150000 0.662587031722069
155000 0.662036657333374
160000 0.6629758477211
165000 0.663124307990074
170000 0.665047690272331
175000 0.666344568133354
180000 0.666421115398407
185000 0.668246537446976
190000 0.666775912046432
195000 0.669402688741684
200000 0.67000487446785
205000 0.667707428336143
210000 0.668522551655769
215000 0.666492208838463
220000 0.668692916631699
225000 0.670354336500168
230000 0.670758634805679
235000 0.671145707368851
};
\addlegendentry{parallel-concat}
\addplot [thick,color2]
table {%
5000 0.609621703624725
10000 0.609266877174377
15000 0.608587637543678
20000 0.608246713876724
25000 0.611010491847992
30000 0.612805336713791
35000 0.61785577237606
40000 0.624222382903099
45000 0.626408353447914
50000 0.631976559758186
55000 0.633373990654945
60000 0.635476991534233
65000 0.636345148086548
70000 0.636688813567162
75000 0.637759283185005
80000 0.63647748529911
85000 0.638708308339119
90000 0.638399988412857
95000 0.639350071549416
100000 0.641250297427177
105000 0.642150819301605
110000 0.646145045757294
115000 0.648251533508301
120000 0.650489509105682
125000 0.651003316044807
130000 0.650979816913605
135000 0.649146974086761
140000 0.6484504789114
145000 0.649817854166031
150000 0.650312840938568
155000 0.653024449944496
160000 0.655553534626961
165000 0.65463761985302
170000 0.653495743870735
175000 0.653668358922005
180000 0.652703821659088
185000 0.652970626950264
190000 0.652960553765297
195000 0.651807889342308
200000 0.651408120989799
205000 0.651009112596512
210000 0.653872579336166
215000 0.653975248336792
220000 0.65466696023941
225000 0.656056985259056
230000 0.65435878932476
235000 0.655195370316505
};
\addlegendentry{cross-sum}
\addplot [line width=2.0pt, color3]
table {%
5000 0.607364505529404
10000 0.606968879699707
15000 0.607783854007721
20000 0.607588171958923
25000 0.60868476331234
30000 0.613596439361572
35000 0.619793236255646
40000 0.627591773867607
45000 0.631772994995117
50000 0.634424537420273
55000 0.634435310959816
60000 0.635495856404305
65000 0.638723760843277
70000 0.641247898340225
75000 0.64394336938858
80000 0.644525468349457
85000 0.645448699593544
90000 0.644832879304886
95000 0.647225946187973
100000 0.646483033895493
105000 0.645708471536636
110000 0.648904830217361
115000 0.65044878423214
120000 0.653215184807777
125000 0.653933003544807
130000 0.653343766927719
135000 0.651272624731064
140000 0.651888534426689
145000 0.652619257569313
150000 0.65303073823452
155000 0.65269248187542
160000 0.651460826396942
165000 0.651540204882622
170000 0.652187079191208
175000 0.653680875897408
180000 0.654272720217705
185000 0.65314906835556
190000 0.652710899710655
195000 0.652027815580368
200000 0.652680888772011
205000 0.656447172164917
210000 0.656094431877136
215000 0.659367755055428
220000 0.659342139959335
225000 0.658961176872253
230000 0.66004441678524
235000 0.658319681882858
};
\addlegendentry{cross-concat}
\addplot [line width=2.0pt, color4]
table {%
5000 0.608646348118782
10000 0.604965761303902
15000 0.601441994309425
20000 0.602763041853905
25000 0.607012465596199
30000 0.612868949770927
35000 0.61718462407589
40000 0.621092110872269
45000 0.622351706027985
50000 0.622246354818344
55000 0.623708158731461
60000 0.62380875647068
65000 0.627416089177132
70000 0.632045775651932
75000 0.635075479745865
80000 0.637971252202988
85000 0.636917307972908
90000 0.63601890206337
95000 0.636704429984093
100000 0.636478334665298
105000 0.637905180454254
110000 0.639311984181404
115000 0.641407489776611
120000 0.644438937306404
125000 0.647825822234154
130000 0.648929089307785
135000 0.64798142015934
140000 0.646652519702911
145000 0.645524844527245
150000 0.644722789525986
155000 0.646567702293396
160000 0.648285210132599
165000 0.651565313339233
170000 0.654553905129433
175000 0.655937641859055
180000 0.655877530574799
185000 0.653390556573868
190000 0.651605099439621
195000 0.650764673948288
200000 0.652319997549057
205000 0.654980719089508
210000 0.656458377838135
215000 0.656901836395264
220000 0.656393095850945
225000 0.655150547623634
230000 0.654573425650597
235000 0.652286380529404
};
\addlegendentry{independent++}
\addplot [thick,color5]
table {%
5000 0.60714054107666
10000 0.607082158327103
15000 0.605858981609344
20000 0.605163633823395
25000 0.604133531451225
30000 0.60532058775425
35000 0.60725836455822
40000 0.61185958981514
45000 0.616917788982391
50000 0.618075788021088
55000 0.621685534715652
60000 0.620797634124756
65000 0.62083038687706
70000 0.624274387955666
75000 0.624823495745659
80000 0.626459583640099
85000 0.627590641379356
90000 0.627424687147141
95000 0.629559308290482
100000 0.629107773303986
105000 0.630091309547424
110000 0.631022244691849
115000 0.628612175583839
120000 0.628956526517868
125000 0.630858823657036
130000 0.629866197705269
135000 0.633931741118431
140000 0.63444696366787
145000 0.632388323545456
150000 0.634294405579567
155000 0.631519228219986
160000 0.632494926452637
165000 0.634151339530945
170000 0.63515242934227
175000 0.636957049369812
180000 0.63673597574234
185000 0.636311933398247
190000 0.634322494268417
195000 0.63375335931778
200000 0.633740663528442
205000 0.634223937988281
210000 0.634928941726685
215000 0.637332111597061
220000 0.639633789658546
225000 0.640514567494392
230000 0.639631673693657
235000 0.638322189450264
};
\addlegendentry{independent}
\end{axis}

\end{tikzpicture}

%% file: dev_bleu.tex
\begin{tikzpicture}

\definecolor{color0}{rgb}{0.117647058823529,0.564705882352941,1}
\definecolor{color1}{rgb}{0.529411764705882,0.807843137254902,0.980392156862745}
\definecolor{color2}{rgb}{0.196078431372549,0.803921568627451,0.196078431372549}
\definecolor{color3}{rgb}{0.564705882352941,0.933333333333333,0.564705882352941}
\definecolor{color4}{rgb}{1,0.549019607843137,0}
\definecolor{color5}{rgb}{1,0.647058823529412,0}

\definecolor{color4}{RGB}{254,97,0}
\definecolor{color5}{RGB}{254,97,0}
\definecolor{color2}{RGB}{17,119,51}
\definecolor{color3}{RGB}{17,119,51}
\definecolor{color0}{RGB}{100,143,255}
\definecolor{color1}{RGB}{100,143,255}

\begin{axis}[
legend cell align={left},
legend style={fill opacity=0.8, draw opacity=1, text opacity=1, at={(0.97,0.03)}, anchor=south east, draw=white!80!black},
tick align=outside,
tick pos=left,
x grid style={white!69.0196078431373!black},
xmin=-0.35, xmax=7.35,
xtick style={color=black},
xtick={-1,0,1,2,3,4,5,6,7,8},
xtick={0,1,2,3,4,5,6,7},
xtick={0,1,2,3,4,5,6,7},
xtick={0,1,2,3,4,5,6,7},
xtick={0,1,2,3,4,5,6,7},
xtick={0,1,2,3,4,5,6,7},
xtick={0,1,2,3,4,5,6,7},
xticklabels={,pt,es,ro,it,fr,nl,ru,de,},
xticklabels={pt,es,ro,it,fr,nl,ru,de},
xticklabels={pt,es,ro,it,fr,nl,ru,de},
xticklabels={pt,es,ro,it,fr,nl,ru,de},
xticklabels={pt,es,ro,it,fr,nl,ru,de},
xticklabels={pt,es,ro,it,fr,nl,ru,de},
xticklabels={pt,es,ro,it,fr,nl,ru,de},
y grid style={white!69.0196078431373!black},
ymin=-1.57, ymax=1.07,
ytick style={color=black},
typeset ticklabels with strut,
xticklabel style={yshift=5pt}
]
\addplot [line width=2.0pt, color0]
table {%
0 0
1 0
2 0
3 0
4 0
5 0
6 0
7 0
};
\addplot [thick, color1]
table {%
0 -1.45
1 -1.12
2 -1.21
3 -0.34
4 -0.530000000000001
5 -0.600000000000001
6 -0.85
7 -0.379999999999999
};
\addplot [line width=2.0pt, color2]
table {%
0 -1
1 -0.530000000000001
2 -0.48
3 -0.0599999999999987
4 -0.00999999999999801
5 0.449999999999999
6 0.420000000000002
7 -0.199999999999999
};
\addplot [thick, color3]
table {%
0 -0.370000000000001
1 -0.360000000000003
2 -0.119999999999997
3 0.0800000000000018
4 0.620000000000001
5 0.469999999999999
6 0.0800000000000001
7 0.109999999999999
};
\addplot [line width=2.0pt, color4]
table {%
0 -0.0499999999999987
1 0.239999999999998
2 0.690000000000001
3 0.75
4 0.640000000000001
5 0.620000000000001
6 0.75
7 0.949999999999999
};
\addplot [thick, color5]
table {%
0 -0.73
1 -0.59
2 -0.0299999999999976
3 0.32
4 0.43
5 0.149999999999999
6 0.200000000000001
7 0.68
};
\end{axis}

\end{tikzpicture}